# Classifying Clinical Outcome of Epilepsy Patients with Ictal Chirp Embeddings

Nooshin Bahador[1], Milad Lankarany[1,2,3,4]

[1] Krembil Research Institute – University Health Network (UHN), Toronto, ON, Canada
[2] Institute of Biomaterials & Biomedical Engineering (IBBME), University of Toronto, Toronto, ON, Canada
[3] Department of Physiology, University of Toronto, Toronto, ON, Canada
[4] KITE Research Institute, Toronto Rehabilitation Institute - University Health Network (UHN), Toronto, ON, Canada
* To whom correspondence should be addressed:
Milad Lankarany (milad.lankarany@uhn.ca)
The Krembil Research Institute – University Health Network (UHN)
60 Leonard Ave, Toronto, M5T 0S8, Canada

**Abstract:**

   This study presents a pipeline leveraging t-Distributed Stochastic Neighbor Embedding (t-SNE) for interpretable visualizations of chirp features across diverse outcome scenarios. The dataset, comprising chirp-based temporal, spectral, and frequency metrics. Using t-SNE, local neighborhood relationships were preserved while addressing the crowding problem through Student t-distribution-based similarity optimization. Three classification tasks were formulated on the 2D t-SNE embeddings: (1) distinguishing clinical success from failure/no-resection, (2) separating high-difficulty from low-difficulty cases, and (3) identifying optimal cases, defined as successful outcomes with minimal clinical difficulty. Four classifiers, namely, Random Forests, Support Vector Machines, Logistic Regression, and k-Nearest Neighbors, were trained and evaluated using stratified 5-fold cross-validation. Across tasks, the Random Forest and k-NN classifiers demonstrated superior performance, achieving up to 88.8% accuracy in optimal case detection (successful outcomes with minimal clinical difficulty). Additionally, feature influence sensitivity maps were generated using SHAP explanations applied to model predicting t-SNE coordinates, revealing spatially localized feature importance within the embedding space. These maps highlighted how specific chirp attributes drive regional clustering and class separation, offering insights into the latent structure of the data. The integrated framework showcases the potential of interpretable embeddings and local feature attribution for clinical stratification and decision support.

# 1. Introduction

Epileptic seizures are the result of abrupt, unregulated bursts of atypical electrical activity in the brain, typically triggered by an imbalance between excitatory and inhibitory neural mechanisms. This disruption elevates the brain's excitability threshold, interfering with both cognitive and physiological function. When such episodes recur without provocation, the condition is classified as epilepsy—a chronic neurological disorder marked by varied symptoms. The causes of seizures are multifactorial, involving hypoxia, genetic predispositions, developmental brain anomalies, and systemic diseases. Clinical presentations differ significantly, affecting sensory experience, motor control, and consciousness, thereby complicating both diagnosis and treatment strategies (Freeman et al., 1993). Electroencephalography (EEG) plays a central role in detecting and analyzing epileptic activity, categorizing it into three primary temporal phases: ictal (during the seizure), interictal (between seizures), and postictal (following seizure cessation). Much of the scientific investigation has concentrated on understanding how seizures initiate, propagate, and resolve during the ictal phase (e.g., Miri et al., 2018; Rich et al., 2020). One proposed mechanism highlights the role of synchronized inhibitory networks, which may suppress normal neuronal activity while elevating extracellular potassium levels, thus creating an environment conducive to seizure initiation (de Curtis and Avoli, 2016). These physiological processes often manifest in EEG recordings as specific time-frequency patterns known as "chirps"—transient bursts exhibiting gradual frequency modulation (Grinenko et al., 2018).

Chirp-like signatures have been extensively reported in intracranial EEG (iEEG) recordings during seizure episodes (Bahador et al., 2024; Bahador et al., 2025; Li et al., 2020; Gnatkovsky et al., 2011; Kurbatova et al., 2016; Sen et al., 2007; Niederhauser et al., 2003; Schiff et al., 2000; Feltane et al., 2013; Gnatkovsky et al., 2019a; Benedetto and Colella, 1995). Recent studies propose that these high-frequency chirps may offer diagnostic insights by highlighting epileptogenic regions (Di Giacomo et al., 2024). However, the temporal emergence of chirps remains contested—some findings link them to pre-onset activity, while others observe them during seizure progression (e.g., Kurbatova et al., 2016; Niederhauser et al., 2003). Importantly, outcomes from surgical resections targeting chirp-generating areas have shown significant improvement, suggesting their potential utility in pre-surgical planning. Di Giacomo et al. (2024) revealed that high-frequency chirps observed in SEEG are reliable indicators of epileptogenic zones in patients with drug-resistant focal epilepsy. These patterns appeared consistently across various seizure types and anatomical contexts and overlapped with clinically determined seizure foci. Furthermore, surgical excision of regions linked to chirps corresponded with favorable postoperative results, underscoring the clinical relevance of chirp analysis in cases lacking clear structural imaging.

Despite encouraging evidence supporting ictal chirps as biomarkers for predicting surgical outcomes, chirp characteristics and its clinical applicability remain underexplored. To address these gaps, we utilize a large, annotated dataset derived from the OpenNeuro Epilepsy-iEEG-Multicenter-Dataset (Li et al., 2022), comprising iEEG recordings from 13 patients across multiple centers. Chirp patterns were semi-automatically annotated using a hybrid method that combines expert-drawn bounding boxes with algorithmic ridge detection and curve fitting to extract quantitative features. We applied dimensionality reduction using t-distributed Stochastic Neighbor Embedding (t-SNE) to map high-dimensional chirp characteristics into a 2D space, enabling visualization of their relationship to clinical variables. These embeddings were then used to evaluate multiple clinical classification tasks, including outcome prediction, case difficulty stratification, and identification of optimal surgical cases. Four classifiers—Random Forest, SVM, Logistic Regression, and k-Nearest Neighbors—were tested, and their performance assessed using cross-validation and standard metrics. To further interpret the embedding structure, SHAP-based sensitivity maps were created, revealing how individual features contribute to the spatial layout of chirp clusters in embedding space.

## 2. Method

### 2.1. Chirp Dataset derived from Intracranial EEG Recordings

This study used iEEG data from the OpenNeuro Epilepsy-iEEG-Multicenter-Dataset, featuring recordings from 13 patients across four centers, organized in BIDS format with clinical metadata (Li et al., 2022). A secondary dataset of 22,721 spectrograms was generated (Bahador & Lankarany, 2025), where chirp patterns were semi-automatically annotated using a hybrid method: users drew bounding boxes, and an algorithm applied ridge detection and model fitting (e.g., linear, exponential) to extract features like start/end time and frequency, duration, RMSE, $R^2$, and direction. Clinical data included seizure onset zones, suspected epileptogenic regions, surgical outcomes (Engel/ILAE), and case difficulty. The annotated dataset and related materials are publicly available on [GitHub](GitHub).

### 2.2. Dimensionality Reduction Using t-SNE

The high-dimensional feature space was transformed into a two-dimensional representation using t-Distributed Stochastic Neighbor Embedding (t-SNE), which is a nonlinear technique for reducing the number of dimensions while preserving the local relationships among data points. Given a set of high-dimensional data points, t-SNE creates a corresponding set of points in a lower-dimensional space—typically two dimensions—that maintains the proximity of similar points as much as possible.

- Step 1: Similarity Calculation in High-Dimensional Space
  The process involves two main steps. First, in the original high-dimensional space, the algorithm calculates how similar each pair of data points is by interpreting these similarities as conditional probabilities. This means the likelihood of one point being a neighbor of another is measured using a Gaussian-based measure, where the width of the Gaussian is automatically adjusted to maintain a certain level of complexity called perplexity. These probabilities are then made symmetric to ensure that the similarity between any two points is equally weighted from both perspectives.
- Step 2: Similarity Calculation and Optimization in Low-Dimensional Space
  Second, in the low-dimensional space, the similarities between points are computed using a distribution known as the student t-distribution with one degree of freedom, which helps to manage the so-called "crowding problem" by allowing distant points to be modeled with heavier tails, ensuring better separation. The algorithm then optimizes the placement of points in this low-dimensional space by minimizing the difference between the high-dimensional and low-dimensional similarity distributions, using an iterative method called gradient descent enhanced with momentum to accelerate convergence.

### 2.3. Feature Weighting and Standardization

Before applying t-SNE, the features could be optionally weighted by multiplying each feature by a corresponding weight and standardized by subtracting the mean and dividing by the standard deviation, which helps to ensure that all features contribute equally to the distance calculations.

### 2.4. Clinical Classification Tasks Using t-SNE Embeddings

Following the dimensionality reduction, three clinical classification tasks were evaluated based on the two-dimensional embeddings generated by t-SNE. These tasks included distinguishing successful outcomes from failures or non-responses, separating cases based on difficulty levels, and identifying optimal cases defined by success and low difficulty.

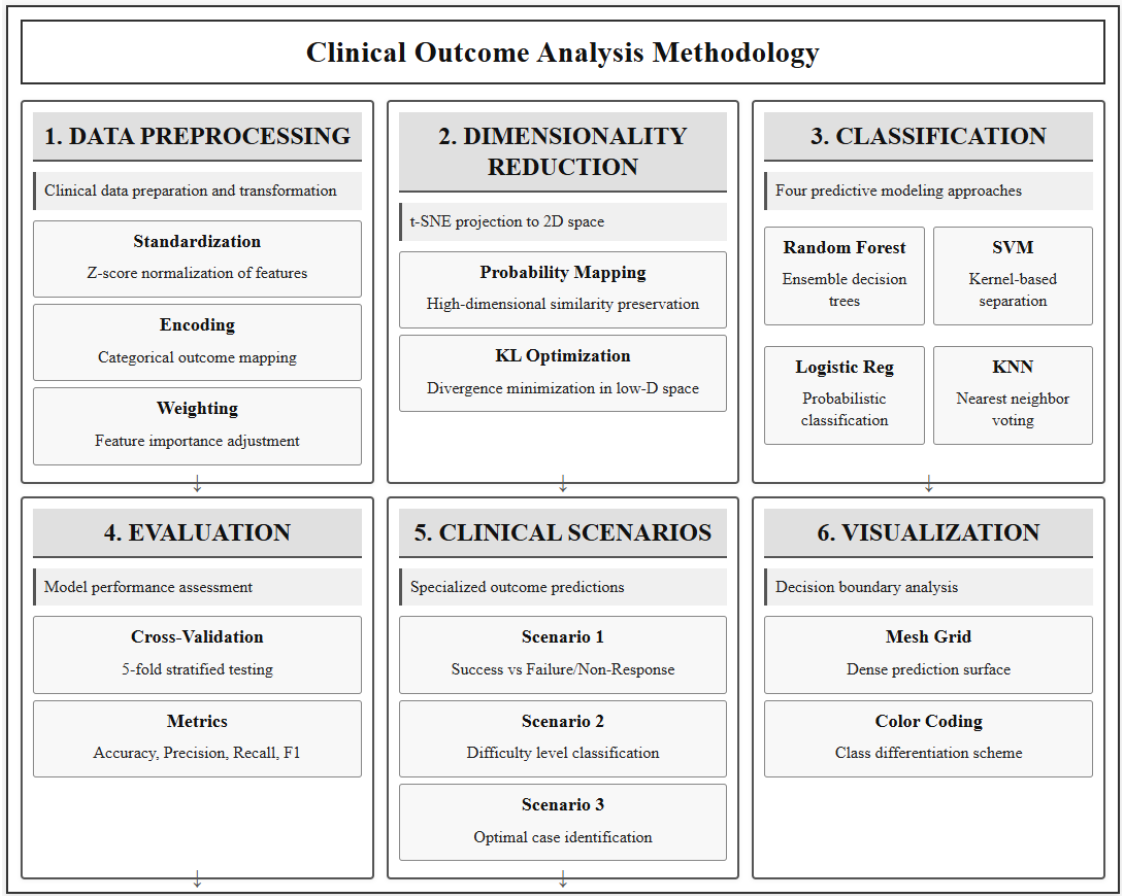

**Fig. 1.** Overview of the t-SNE Clinical Outcome Analysis Pipeline. A structured six-step workflow illustrating the application of t-distributed Stochastic Neighbor Embedding (t-SNE) for clinical data analysis. Step 1 involves preprocessing, including feature scaling and weighting. Step 2 applies t-SNE for nonlinear dimensionality reduction to 2D. Step 3 integrates four classifiers—Random Forest, Support Vector Machine, Logistic Regression, and K-Nearest Neighbors—for predictive modeling. Step 4 assesses performance using cross-validation and stratified train-test splits. Step 5 defines clinical prediction scenarios based on symptom presence and severity. Step 6 visualizes classifier decision boundaries over t-SNE-reduced feature space, highlighting classification zones with class-specific color encoding. Detailed methodology is provided in supplementary **Figs. Supp. 1.1–1.6** in Appendix.

## 2.5. Classifiers Tested on the Embeddings

Four different classifiers were tested in each scenario. The first was a Random Forest, which builds many decision trees and splits data based on a measure of impurity called Gini impurity, reflecting how mixed the classes are at each decision point (Breiman, 2001). The second was a Support Vector Machine (SVM), which finds the best boundary that separates classes by solving an optimization problem balancing margin size and classification errors (Cortes & Vapnik, 1995). The third was Logistic Regression, a probabilistic model estimating the likelihood of class membership based on a logistic function applied to a weighted sum of features (Cox, 1958). The fourth was the k-Nearest Neighbors method, which assigns class labels based on the majority vote among the closest points in the embedding space, measured using the usual geometric distance (Cover & Hart, 1967).

## 2.6. Performance Metrics for Model Evaluation

Model performance was evaluated using standard metrics including accuracy (the proportion of correctly classified samples), precision (the fraction of positive predictions that were correct), recall (the fraction of actual positives correctly identified), and the F1-score, which balances precision and recall into a single measure.

## 2.7. Cross-Validation Strategy

To ensure robust evaluation, stratified 5-fold cross-validation was used, meaning the dataset was split into five parts with each fold maintaining the same class proportions as the full dataset. The models were trained and tested across these folds without overlap, providing a reliable estimate of performance.

Fig.1 presents an overview of the t-SNE analysis pipeline, with detailed methodological steps available in Supplementary Fig.1.1 to 1.6 in the Appendix.

## 3. Results

### 3.1. Data Characteristics and t-SNE Visualization

The clean dataset comprised 20,210 samples including three key chirp features: temporal duration, frequency onset, and spectral duration (Bahador & Lankarany, 2025). The outcome distribution showed 28.2% success (S), 30.1% no resection (NR), and 41.7% failure (F) cases (Fig.2). Clinical difficulty was distributed to four levels: 17.6% level 1 (very easy), 8.0% level 2 (easy), 28.9% level 3 (difficult), and 45.5% level 4 (very difficult) cases (Fig.2).

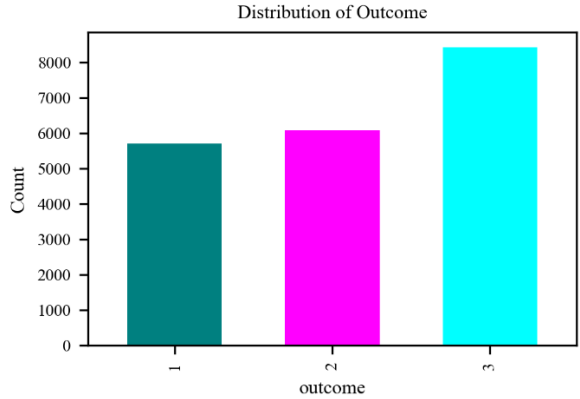
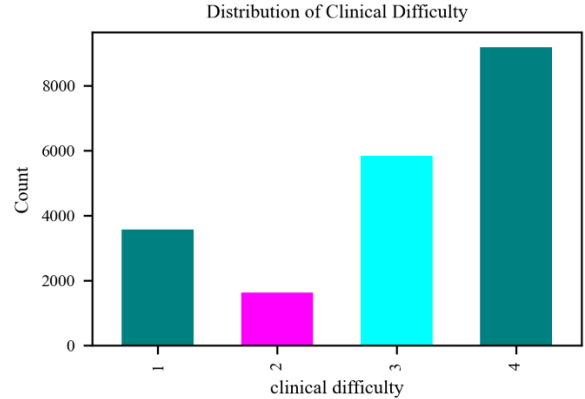

**Fig.2.** Distribution of clinical outcomes in the dataset (S=Success, NR=No Resection, F=Failure).

**Fig.3.** Distribution of clinical difficulty levels in the dataset.

### 3.2. t-SNE visualization of the feature space

t-SNE visualization of the feature space (n=1,549 samples after preprocessing) revealed distinct patterns when colored by clinical outcomes (Fig.4) and difficulty levels (Fig.5). The equal weights scenario ([1,1,1]) showed moderate separation between classes, while weighted scenarios ([2,1,1], [1,2,1], [1,1,2]) demonstrated varying degrees of improved separation.

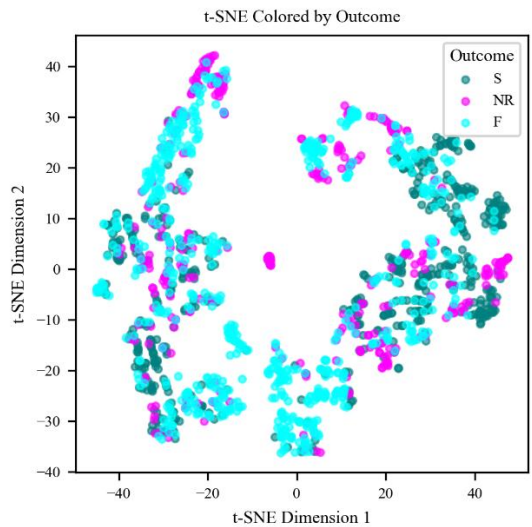
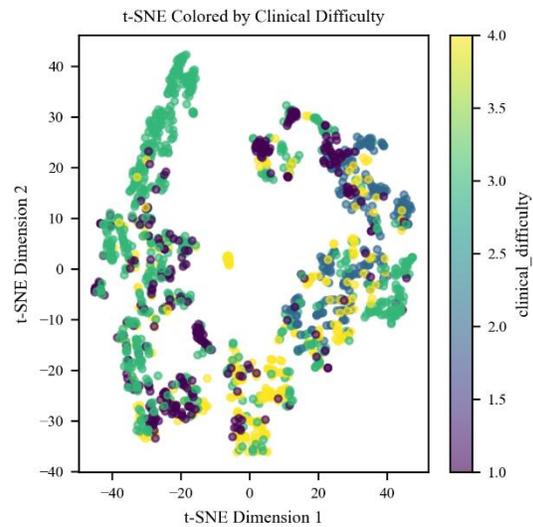

**Fig.4**. t-SNE embedding colored by clinical outcome (S=teal, NR=magenta, F=cyan).

**Fig.5**. t-SNE embedding colored by clinical difficulty level (viridis colormap).

## 3.3. Classification Performance

Three classification scenarios (Fig.6) were evaluated: (1) distinguishing success from failure/no resection; (2) classifying high vs. low difficulty cases; and (3) identifying optimal cases (success with low difficulty).

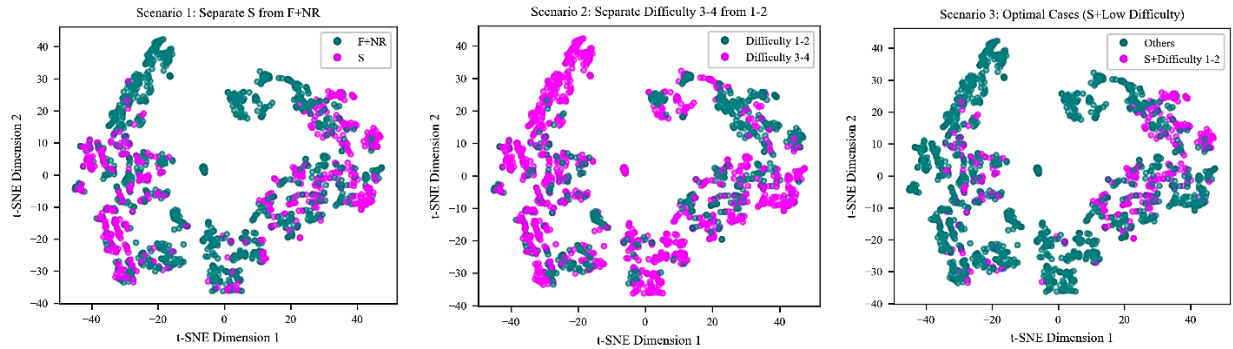

**Fig.6. Classification performance for three scenarios:** (left) distinguishing successful outcomes from failure or no resection; (middle) classifying high vs. low difficulty cases; and (right) identifying optimal cases (i.e., success with low difficulty).

The following subsections present decision boundary visualizations for each classifier—Random Forest, SVM, Logistic Regression, and K-Nearest Neighbors—which illustrate how each algorithm partitions the t-SNE embedding space. Random Forest generates complex, non-linear boundaries by aggregating multiple decision trees. SVM defines a hyperplane that maximizes the margin between classes (Cortes & Vapnik, 1995). Logistic Regression produces a linear boundary using sigmoid-based probability outputs (Hosmer et al., 2013), while KNN creates irregular boundaries based on local neighbor voting (Cover et al., 1967). Confusion matrices summarize classification performance by displaying true positives, false positives, true negatives, and false negatives (Provost & Fawcett, 2013). Performance metrics also provide further evaluation: Accuracy: Proportion of correctly classified instances (Powers, 2020). Precision: Proportion of positive predictions that are correct (Sokolova & Lapalme, 2009). Recall (Sensitivity): Proportion of actual positives correctly identified (Fawcett, 2006). F1-Score: Harmonic mean of precision and recall, balancing both (Chinchor, 1992).

### 3.3.1. Scenario 1- Separating Success from Failure/No Resection

Random Forest classifier achieved the highest accuracy (78.7% ± 2.3%) in distinguishing success (S) from combined failure/No Resection (F+NR) cases, with precision=0.700, recall=0.660, and F1=0.680 (Fig.7). K-Nearest Neighbors performed comparably (77.6% ± 2.9% accuracy), while SVM and Logistic Regression showed lower performance (71.2% and 67.7% accuracy respectively).

### 3.3.2. Scenario 2- Separating High vs Low Difficulty Cases

Classification between difficulty levels 3-4 versus 1-2 showed improved performance, with K-Nearest Neighbors achieving 82.8% ± 0.9% accuracy (precision=0.850, recall=0.890, F1=0.870)

(Fig.8). Random Forest followed closely (81.5% ± 0.8% accuracy), demonstrating robust separation between difficulty groups.

**3.3.3. Scenario 3-** Identifying Optimal Cases (Success with Low Difficulty)

The most distinct separation occurred when identifying optimal cases (S with difficulty 1-2) versus all others. K-Nearest Neighbors achieved 88.8% ± 1.2% accuracy (precision=0.755, recall=0.710, F1=0.732), followed closely by Random Forest (88.6% ± 1.3% accuracy) (Fig.9). The decision boundary visualization clearly shows the separation between optimal cases (magenta) and other outcomes (teal).

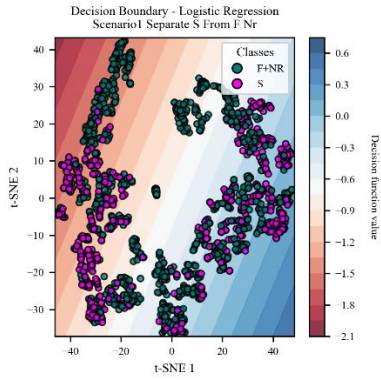
**(A)** Logistic Regression Decision Boundary

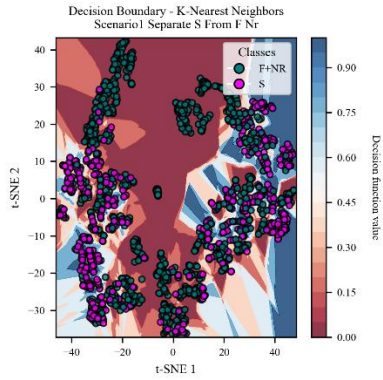
**(B)** Nearest Neighbors Decision Boundary

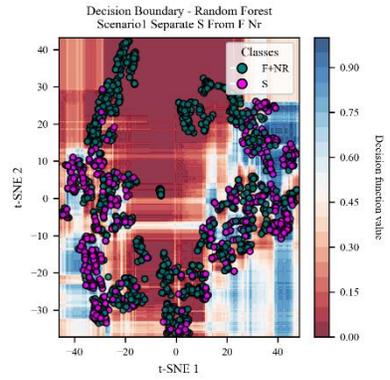
**(C)** Random Forest Decision Boundary

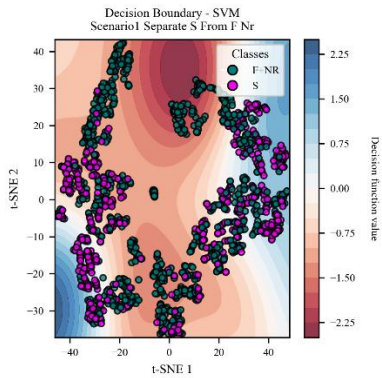
**(D)** SVM Decision Boundary

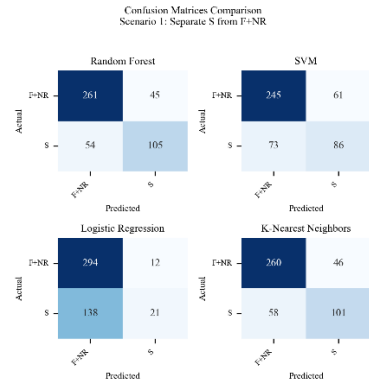
**(E)** Confusion Matrices Comparison

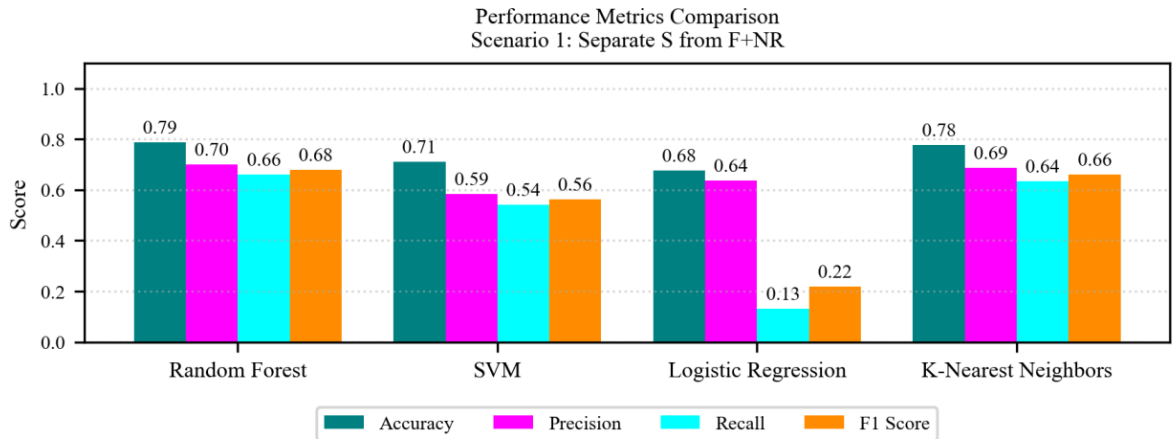
**(F)** Performance metrics comparison for Scenario 1 classification.
**Fig.7.** Scenario 1 classification: separating S from F and NR cases

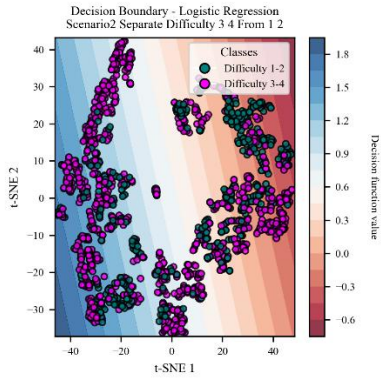

**(A)** Logistic Regression Decision Boundary

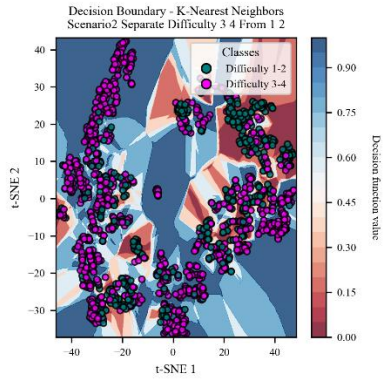

**(B)** Nearest Neighbors Decision Boundary

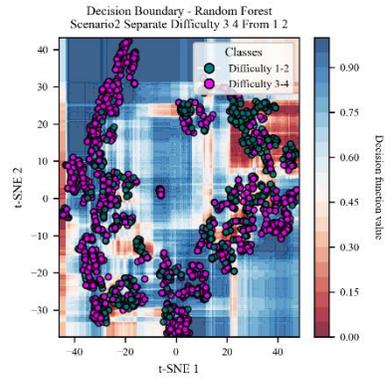

**(C)** Random Forest Decision Boundary

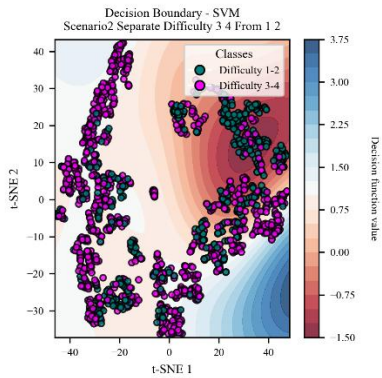

**(D)** SVM Decision Boundary

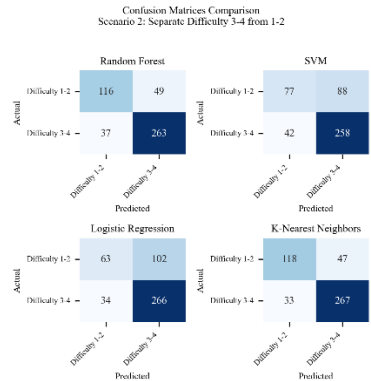

**(E)** Confusion Matrices Comparison

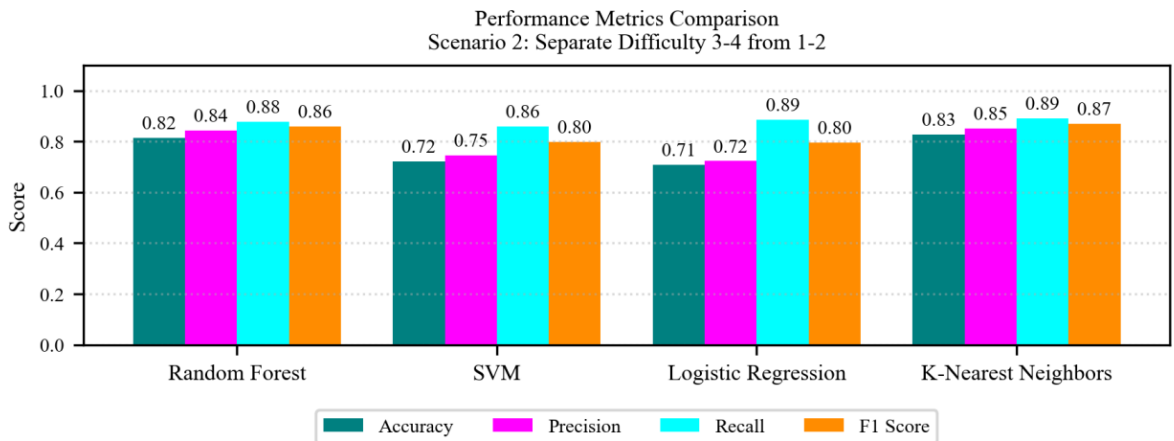

**(F)** Performance metrics comparison for Scenario 2 classification.

**Fig.8.** Scenario 2 classification: separating difficulty 3&4 from 1&2 cases

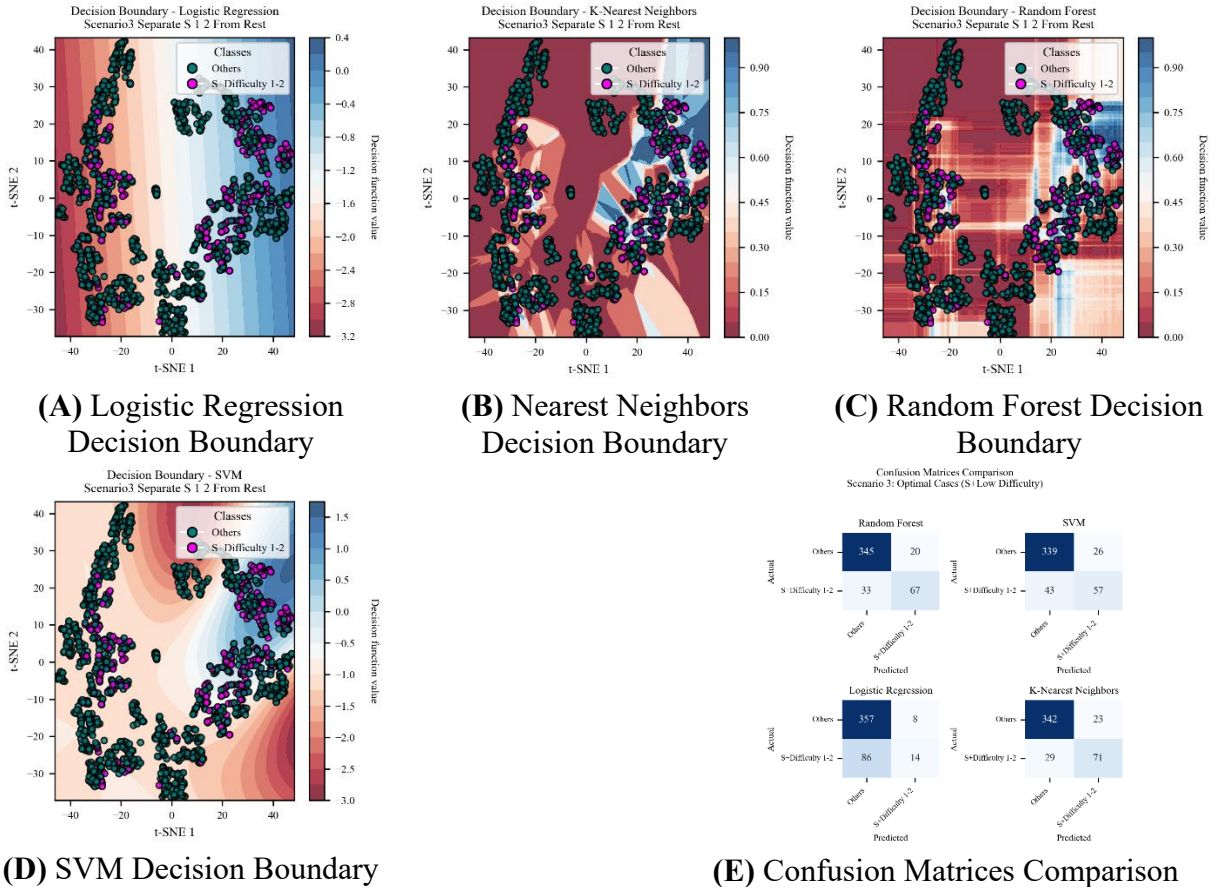

**(A)** Logistic Regression Decision Boundary
**(B)** Nearest Neighbors Decision Boundary
**(C)** Random Forest Decision Boundary
**(D)** SVM Decision Boundary
**(E)** Confusion Matrices Comparison

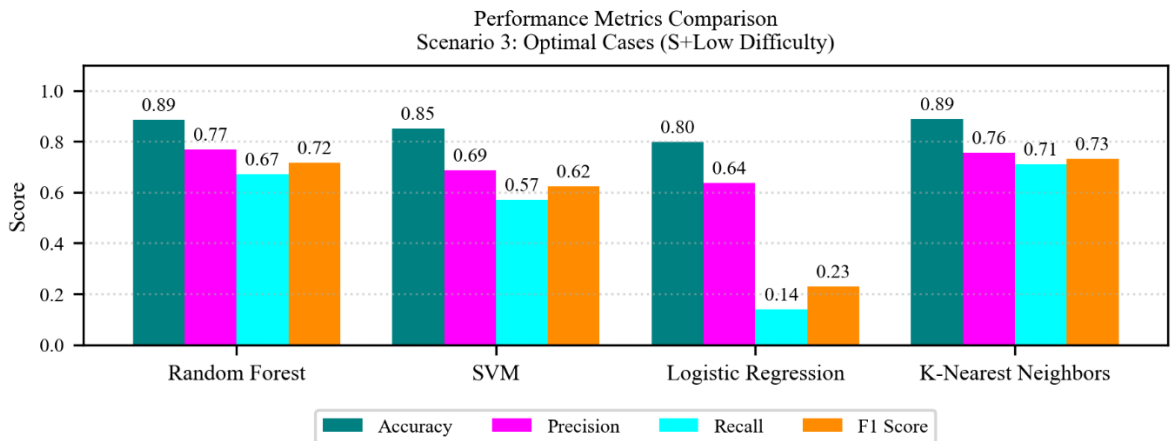

**(F)** Performance metrics comparison for Scenario 3 classification.
**Fig.9.** Scenario 3 classification: separating S with difficulty 1&2 cases from rest

### 3.4. Visualizing Feature Influence in Embedding Space Using Sensitivity Maps

The sensitivity maps (Fig.10) were created to show how much each input feature influences the structure of the t-SNE embedding. To do this, machine learning models (Random Forests) were trained to predict the position of each data point in the 2D t-SNE space based on the input features.

Then, SHAP values were used to explain how important each feature was in determining those positions. For each feature, the influence it had on both dimensions of the t-SNE space was combined into a single value that reflects its overall impact at each point. These values were then visualized by coloring the t-SNE plot, where brighter or darker areas represent stronger or weaker influence from that feature.

The most notable observation in Fig.10 is that maps (A) and (C) show nearly opposite patterns. In the Temporal Duration map (A), the upper-left region shows high sensitivity (yellow-green colors) while the upper-right regions are less sensitive (blue-purple). Conversely, the Spectral Duration map (C) shows the reverse pattern - high sensitivity in the upper-right regions, with lower sensitivity in the upper-left area.

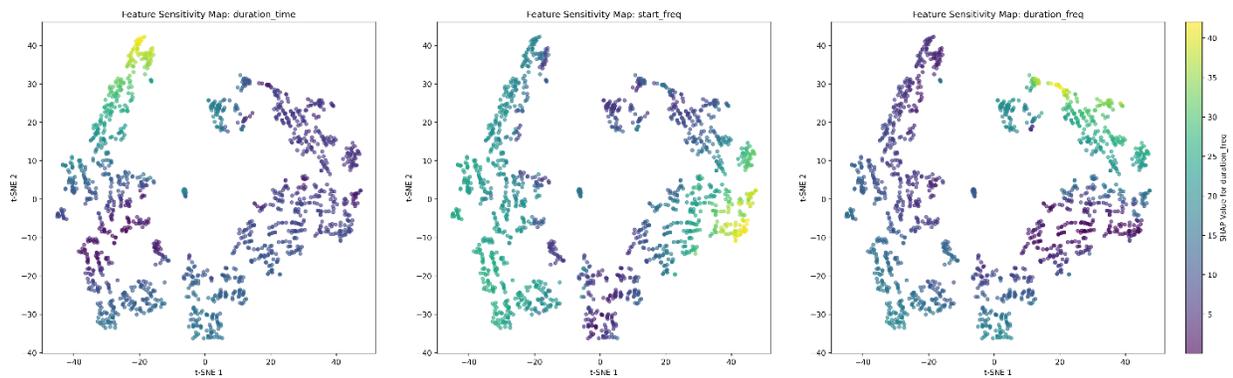

**(A)** Temporal Duration  **(B)** Frequency Onset  **(C)** Spectral Duration

**Fig.10.** Feature Sensitivity in t-SNE Embeddings: SHAP values mapped onto the t-SNE space, where color intensity reflects the magnitude of each feature's contribution. Brighter regions indicate stronger feature influence on the projection. These maps help visualize where each feature has the greatest impact within the embedding space (i.e., their local importance), and identify which features are driving the separation of data points in specific regions. They can reveal clusters where certain features are particularly influential, as well as distinct regions of the embedding space dominated by different features. To support further validation, a straightforward example of feature weighting is presented in Supplementary Fig.10 of the Appendix (**Fig. Supp. 10**).

## 4. Discussion

This study presents a pipeline leveraging chirp features extracted from intracranial EEG (iEEG) recordings to stratify clinical outcomes and case difficulty in patients undergoing epilepsy surgery. By integrating semi-automated chirp annotation, handcrafted feature extraction, nonlinear dimensionality reduction (t-SNE), and a suite of machine learning classifiers, we demonstrate the feasibility of using interpretable, chirp-derived biomarkers for outcome prediction in complex epilepsy cases.

### 4.1. Dimensionality Reduction for Visualization and Stratification

The use of t-SNE enabled a two-dimensional visualization of a high-dimensional feature space while maintaining the intrinsic structure necessary for local class separability. Although t-SNE is non-parametric and unsupervised, it effectively revealed latent groupings corresponding to clinical outcome (S vs. F/NR) and case difficulty (Levels 1–4). Features such as temporal duration, frequency onset, and spectral span of chirps were shown to have nonuniform influence across the t-SNE embedding space, suggesting they encode localized information relevant to clinical phenotypes.

### 4.2. Predictive Modeling on Embedded Feature Space

The classification analyses demonstrated that clinical categories, particularly optimal cases (defined as successful outcome and low difficulty), can be identified with high accuracy in the t-SNE-reduced space. Among classifiers, K-Nearest Neighbors and Random Forests consistently outperformed SVM and Logistic Regression across all scenarios. The superior performance of non-parametric models (KNN, RF) is likely due to their ability to exploit the local neighborhood structure preserved in the t-SNE space, whereas models assuming linearity (Logistic Regression) or requiring separable boundaries (SVM) were less adaptable.

Scenario 3 (identifying optimal cases) achieved the highest classification accuracy (88.8%), underscoring the utility of chirp-derived features and t-SNE embeddings in discriminating clinically favorable profiles. From a translational standpoint, this scenario offers potential for pre-surgical stratification—flagging candidates likely to benefit from surgery with lower complexity and guiding more tailored interventions in high-risk cases.

### 4.3. Feature Interpretability and Clinical Relevance

Interpretability remains a central concern in clinical AI applications. Here, SHAP-based sensitivity maps provided a means to relate chirp features to spatial regions in the 2D embedding, thereby attributing clinical meaning to model outputs. Such visualizations could inform clinicians where specific features (e.g., high-frequency onset) might signify greater or lesser resectability, offering an interpretable bridge between signal properties and outcome likelihoods.

### 4.4. Limitations and Future Work

While promising, several limitations should be acknowledged. First, t-SNE is inherently stochastic and sensitive to hyperparameter settings such as perplexity. Second, the semi-automated annotation pipeline, though efficient, still relies on manual bounding box input, introducing potential bias. Lastly, the limited size of the chirp sample (n=1,549) within the original dataset (n=22,721) reflects constraints that may affect statistical power. Expansion of the dataset and augmentation strategies (e.g., synthetic chirps or time-frequency-domain perturbation) could be beneficial.

**Acknowledgments**

The authors gratefully acknowledge the financial support of the Canadian Neuroanalytics Scholars Program and The Hilary & Galen Weston Foundation. We also recognize the valuable support provided by Campus Alberta Neuroscience, the Hotchkiss Brain Institute at the University of Calgary, the Ontario Brain Institute, and the Neuro at McGill University. Their contributions were essential to the advancement of this research project.

**References**

Adam Li and Sara Inati and Kareem Zaghloul and Nathan Crone and William Anderson and Emily Johnson and Iahn Cajigas and Damian Brusko and Jonathan Jagid and Angel Claudio and Andres Kanner and Jennifer Hopp and Stephanie Chen and Jennifer Haagensen and Sridevi Sarma (2023). Epilepsy-iEEG-Multicenter-Dataset. OpenNeuro. [Dataset] doi: doi:10.18112/openneuro.ds003029.v1.0.6.

Bahador, N., & Lankarany, M. (2025). Semi-automated detection, annotation, and prognostic assessment of ictal chirps in intracranial EEG from patients with epilepsy. doi:10.1101/2025.08.13.670167.

Bahador, N., Lankarany, M. (2025). Chirp Localization via Fine-Tuned Transformer Model: A Proof-of-Concept Study. arXiv preprint arXiv:2503.22713.

Bahador, N., Skinner, F., Zhang, L., & Lankarany, M. (2024). Ictal-related chirp as a biomarker for monitoring seizure progression. doi:10.1101/2024.10.29.620811.

Benedetto, J. J. and Colella, D. (1995). Wavelet analysis of spectrogram seizure chirps. In Wavelet Applications in Signal and Image Processing III, volume 2569, pages 512–521. SPIE.

Breiman, L. (2001). Random forests. Machine Learning, 45(1), 5–32. https://doi.org/10.1023/A:1010933404324.

Chinchor, N. (1992). MUC-4 evaluation metrics. In Proceedings of the 4th Message Understanding Conference (MUC-4), 22-29. Association for Computational Linguistics. https://doi.org/10.3115/1072064.1072067.

Cortes, C., & Vapnik, V. (1995). Support-vector networks. Machine Learning, 20(3), 273–297. https://doi.org/10.1007/BF00994018.

Cover, T., & Hart, P. (1967). Nearest neighbor pattern classification. IEEE Transactions on Information Theory, 13(1), 21–27. https://doi.org/10.1109/TIT.1967.1053964.

Cox, D. R. (1958). The regression analysis of binary sequences. Journal of the Royal Statistical Society: Series B (Methodological), 20(2), 215–242. https://doi.org/10.1111/j.2517-6161.1958.tb00292.x.

de Curtis, M. and Avoli, M. (2016). GABAergic networks jump-start focal seizures. Epilepsia, 57(5):679–687.

Di Giacomo, R., Burini, A., Chiarello, D., Pelliccia, V., Deleo, F., Garbelli, R., De Curtis, M., Tassi, L., & Gnatkovsky, V. (2024). Ictal fast activity chirps as markers of the epileptogenic zone. Epilepsia, 65(6). https://doi.org/10.1111/epi.17995.

Fawcett, T. (2006). An introduction to ROC analysis. Pattern Recognition Letters, 27(8), 861-874. https://doi.org/10.1016/j.patrec.2005.10.010.

Feltane, A., Bartels, G. B., Boudria, Y., and Besio, W. (2013). Analyzing the presence of chirp signals in the electroencephalogram during seizure using the reassignment time-frequency representation and the hough transform. In 2013 6th International IEEE/EMBS Conference on Neural Engineering (NER), pages 186–189. IEEE.

Freeman, J. M., Vining, E. P., and J., P. D. (1993). Seizures and epilepsy in childhood: a guide for parents.

Frosz, M. H. and Andersen, P. E. (2007). Can pulse broadening be stopped? Nature Photonics, 1(11):611–612.

Gnatkovsky, V., Francione, S., Cardinale, F., Mai, R., Tassi, L., Lo Russo, G., and De Curtis, M. (2011). Identification of reproducible ictal patterns based on quantified frequency analysis of intracranial eeg signals. Epilepsia, 52(3):477–488.

Gnatkovsky, V., Pelliccia, V., de Curtis, M., and Tassi, L. (2019a). Two main focal seizure patterns revealed by intracerebral electroencephalographic biomarker analysis. Epilepsia, 60(1):96–106.

Grinenko, O., Li, J., Mosher, J. C., Wang, I. Z., Bulacio, J. C., Gonzalez-Martinez, J., Nair, D., Najm, I., Leahy, R. M., and Chauvel, P. (2018). A fingerprint of the epileptogenic zone in human epilepsies. Brain, 141(1):117–131.

Hosmer, D. W., Jr., Lemeshow, S., & Sturdivant, R. X. (2013). Applied Logistic Regression (3rd ed.). Wiley. DOI: 10.1002/9781118548387.

Kurbatova, P., Wendling, F., Kaminska, A., Rosati, A., Nabbout, R., Guerrini, R., Dulac, O., Pons, G., Cornu, C., Nony, P., et al. (2016). Dynamic changes of depolarizing gaba in a computational model of epileptogenic brain: Insight for dravet syndrome. Experimental Neurology, 283:57–72.

Li, J., Grinenko, O., Mosher, J. C., Gonzalez-Martinez, J., Leahy, R. M., and Chauvel, P. (2020). Learning to define an electrical biomarker of the epileptogenic zone. Human Brain Mapping, 41(2):429–441.

Miri, M. L., Vinck, M., Pant, R., and Cardin, J. A. (2018). Altered hippocampal interneuron activity precedes ictal onset. eLife, 7:e40750.

Niederhauser, J. J., Esteller, R., Echauz, J., Vachtsevanos, G., and Litt, B. (2003). Detection of seizure precursors from depth-eeg using a sign periodogram transform. IEEE Transactions on Biomedical Engineering, 50(4):449–458.

Powers, D. M. W. (2020). Evaluation: From Precision, Recall and F-Measure to ROC, Informedness, Markedness & Correlation. Journal of Machine Learning Technologies, 2(1), 37–63. Preprint: arXiv:2010.16061

Provost, F., & Fawcett, T. (2013). Data Science for Business: What You Need to Know about Data Mining and Data-Analytic Thinking. O'Reilly Media. https://doi.org/10.5555/2533371.

Rich, S., Chameh, H. M., Rafiee, M., Ferguson, K., Skinner, F. K., and Valiante, T. A. (2020). Inhibitory Network Bistability Explains Increased Interneuronal Activity Prior to Seizure Onset. Frontiers in Neural Circuits, 13.

Schiff, S. J., Colella, D., Jacyna, G. M., Hughes, E., Creekmore, J. W., Marshall, A., Bozek-Kuzmicki, M., Benke, G., Gaillard, W. D., Conry, J., et al. (2000). Brain chirps: spectrographic signatures of epileptic seizures. Clinical Neurophysiology, 111(6):953–958.

Sen, A., Kubek, M., and Shannon, H. (2007). Analysis of seizure eeg in kindled epileptic rats. Computational and Mathematical Methods in Medicine, 8(4):225–234.

Wieser, H. G., Blume, W. T., Fish, D., Goldensohn, E., Hufnagel, A., King, D., Lüders, H. (2001). Proposal for a new classification of outcome with respect to epileptic seizures following epilepsy surgery. Epilepsia, 42(s2), 282–286. doi:10.1046/j.1528-1157.2001.4220282.x.

# Appendix

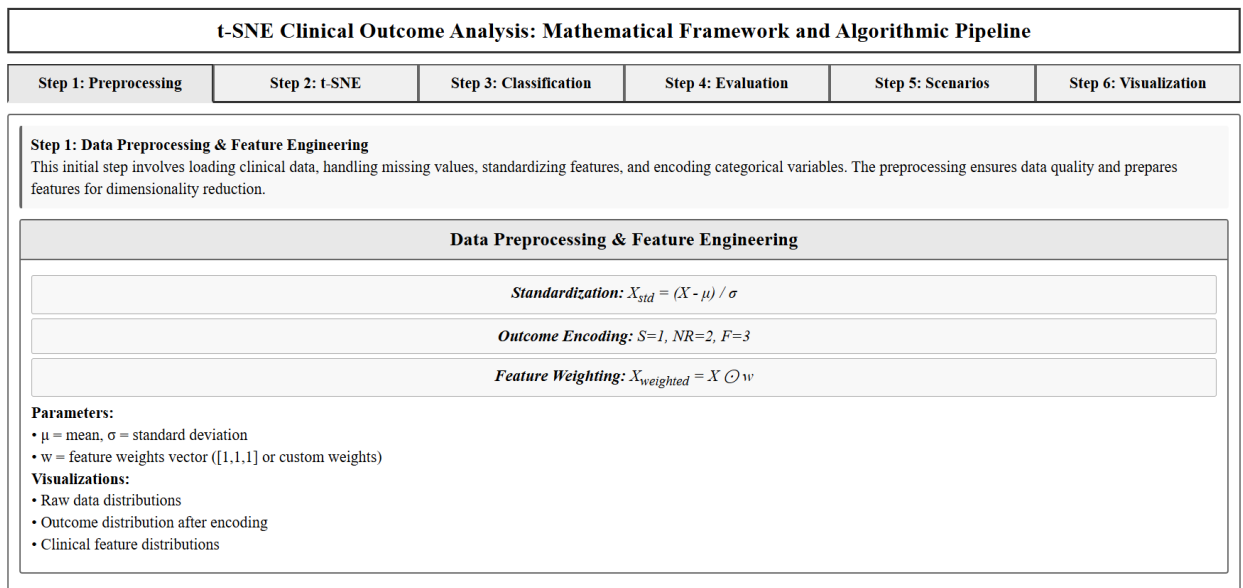

**Fig. Supp. 1.1. Clinical Data Preprocessing and Feature Engineering**

Illustration of the preprocessing pipeline for clinical datasets, including handling of missing values, standardization of continuous features, encoding of categorical outcomes, and application of optional feature weighting. Visual outputs include raw data distributions, outcome encoding maps, and processed feature histograms, preparing the dataset for dimensionality reduction techniques.

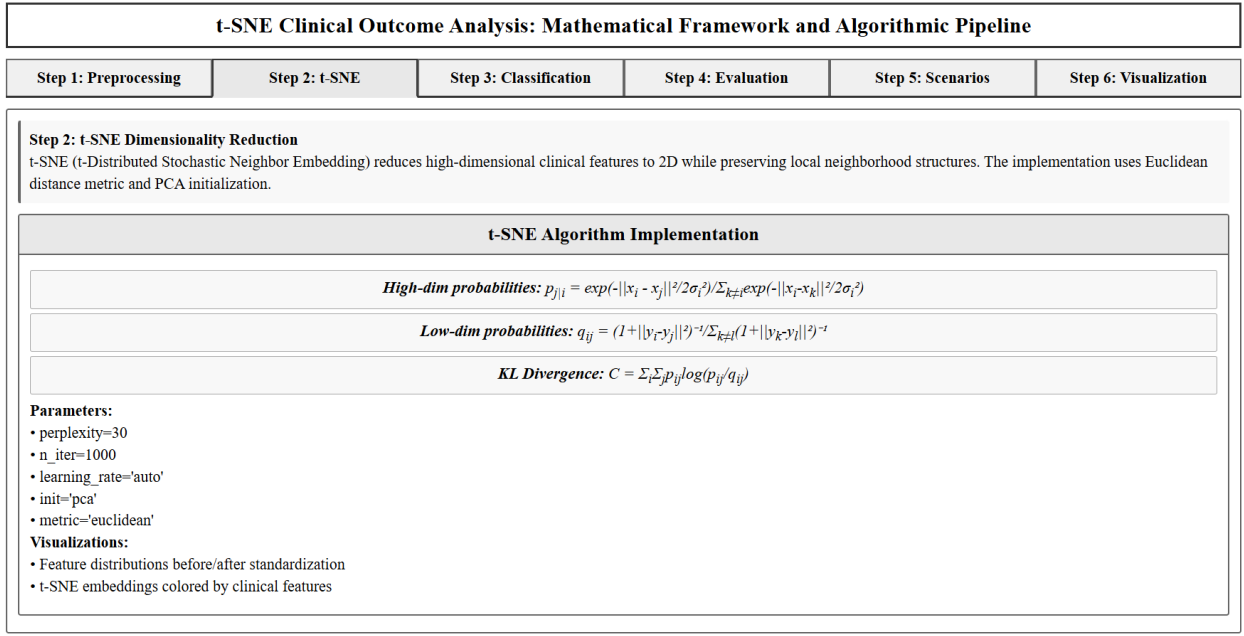

**Fig. Supp. 1.2. t-SNE Dimensionality Reduction of Clinical Features**

Two-dimensional embedding of standardized clinical data using the t-SNE algorithm with PCA initialization. The algorithm preserves local neighborhood structures through probabilistic modeling of pairwise similarities in both high-dimensional and low-dimensional spaces. Key visualizations include t-SNE plots colored by clinical labels and distribution comparisons pre- and post-standardization.

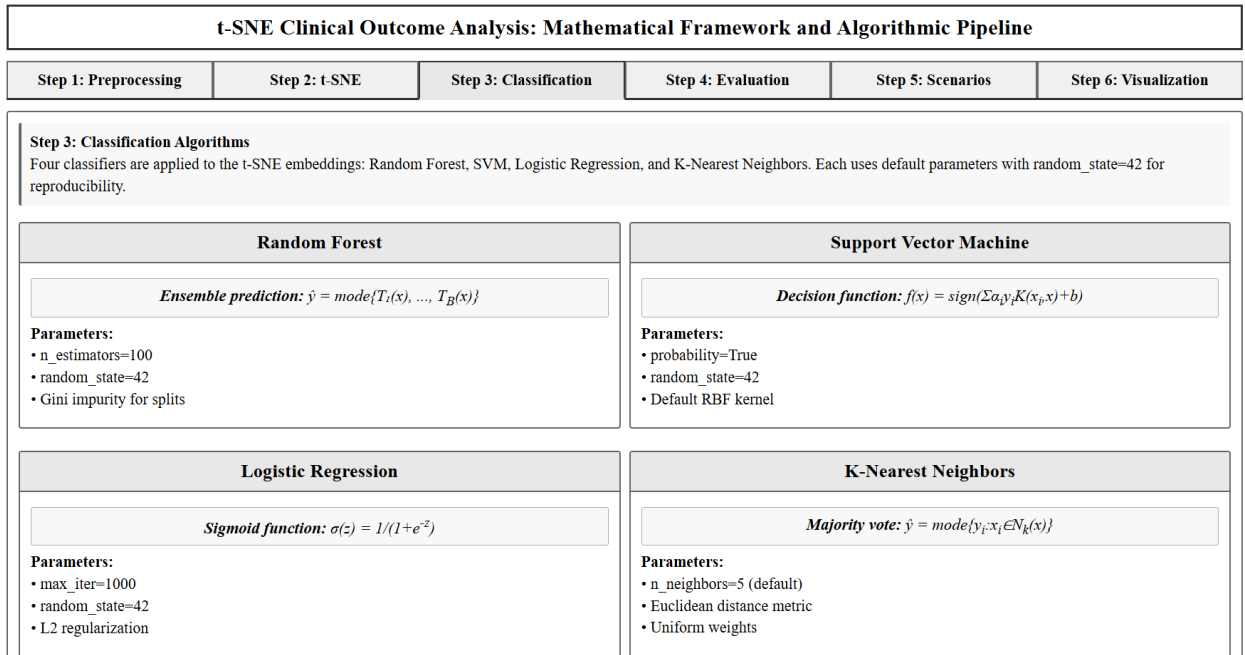

**Fig. Supp. 1.3. Classification Algorithms Applied to t-SNE Embeddings**

Comparison of four classification algorithms—Random Forest, Support Vector Machine, Logistic Regression, and K-Nearest Neighbors—trained on t-SNE embeddings of clinical data. Each classifier's architecture, mathematical formulation, and parameter configuration are presented, with visual analysis supported by model-specific performance overlays on the embedded space.

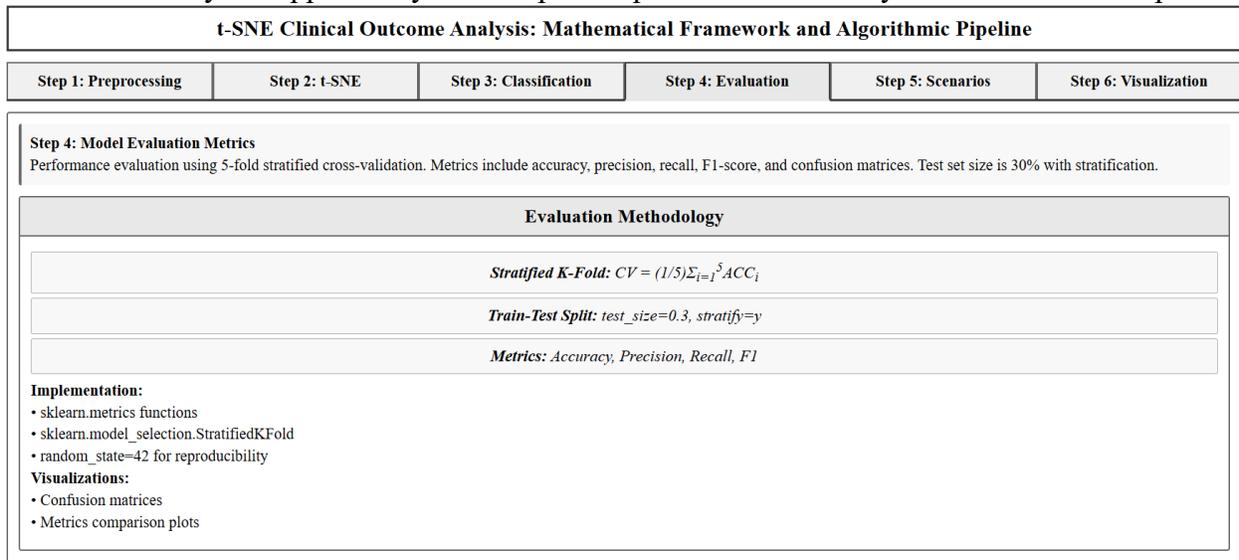

**Fig. Supp. 1.4. Evaluation Metrics and Cross-Validation Strategy**

Summary of the evaluation framework using stratified 5-fold cross-validation and a hold-out test set (30%) for assessing model performance. Performance metrics include accuracy, precision, recall, and F1-score, with results presented via confusion matrices and comparative metric plots for each classifier.

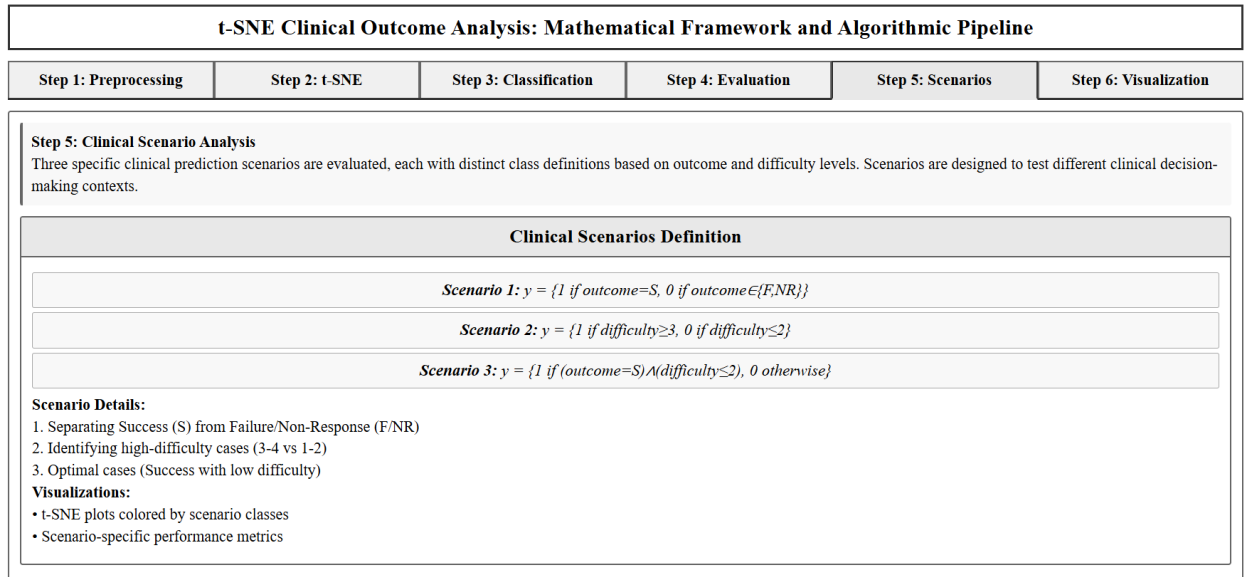

**Fig. Supp. 1.5. Scenario-Based Clinical Outcome Prediction**

Definition and implementation of three clinical prediction scenarios designed to reflect real-world decision-making tasks. Scenarios involve binary classification based on treatment outcome, case difficulty, and a combination of both. Visualizations display t-SNE embeddings annotated by scenario-specific class labels and classifier performance for each case.

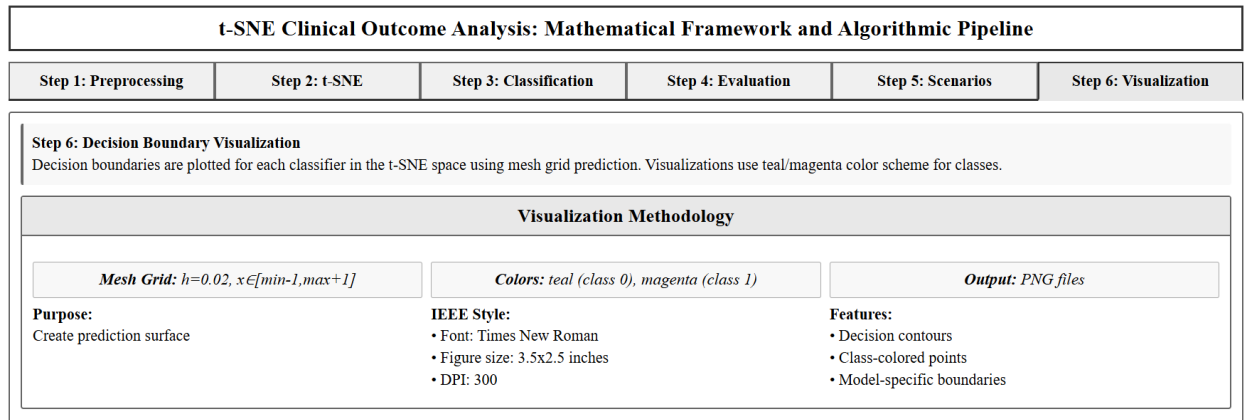

**Fig. Supp. 1.6. Classifier Decision Boundaries in Embedded Feature Space**

Decision boundary visualizations for all classifiers projected onto the t-SNE space using mesh grid predictions. Class regions are color-coded (teal for class 0 and magenta for class 1), with contours indicating model-specific separability. All visualizations follow IEEE publication standards for resolution, font, and layout.

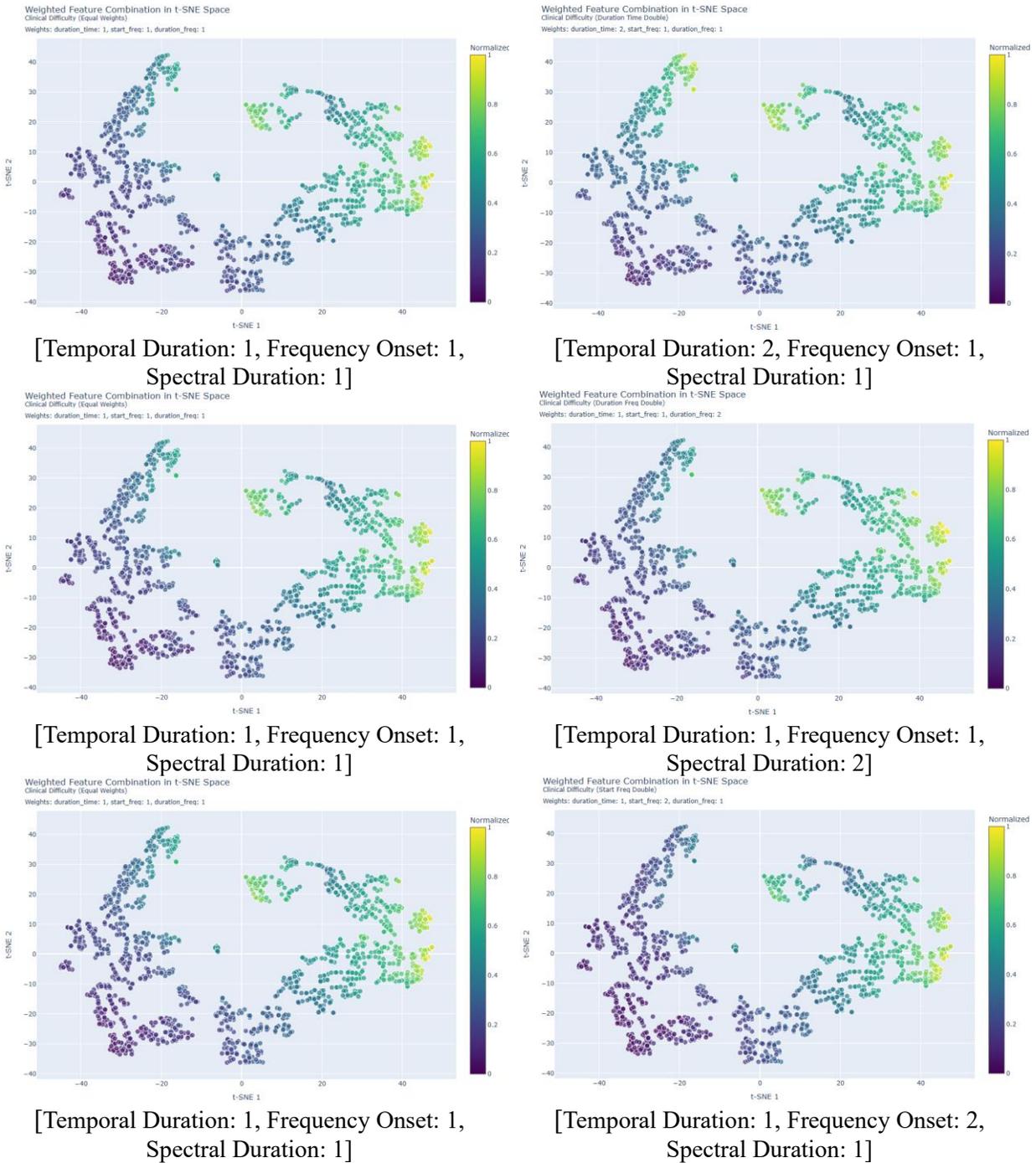

**Fig. Supp. 10.** t-SNE embeddings illustrating the effects of different feature weightings on clinical outcome separation. All panels display 2D embeddings of composite features derived from temporal duration, frequency onset, and spectral duration.

- Left column (baseline): All three features are equally weighted (1:1:1). Colors represent normalized composite scores, reflecting equal contributions from temporal and spectral features. This serves as the baseline for comparison.

- Top right: Temporal duration is double-weighted (2:1:1), emphasizing the influence of event duration. Warmer colors different from baseline indicate regions where longer temporal events may correlate more strongly with clinical outcomes.
- Middle right: Spectral duration is double-weighted (1:1:2), highlighting the role of spectral width.
- Bottom right: frequency onset is double-weighted (1:2:1), emphasizing the initial frequency onset of events.